\documentclass[sigconf]{acmart}

\usepackage{colortbl}
\usepackage[normalem]{ulem}
\useunder{\uline}{\ul}{}
\usepackage[ruled]{algorithm2e}
\usepackage{subcaption}
\usepackage{tcolorbox}
\definecolor{ylp_color1}{RGB}{255,193,193}
\definecolor{ylp_color2}{RGB}{255,228,225}
\usepackage{tikz}
\definecolor{table_color1}{HTML}{EFEFEF}
\definecolor{table_color2}{HTML}{9B9B9B}
\definecolor{table_color3}{HTML}{C0C0C0}
\definecolor{table_edge}{RGB}{0,0,0}
\newtcbox{\mybox}[1][red]{on line, colback = {RGB}{255,228,225}, colframe = {RGB}{255,193,193},  arc=1mm, auto outer arc, boxrule=0.5pt,}
\usepackage{multirow}
\usepackage{booktabs}
\usepackage{wrapfig}
\usepackage{amsfonts}
\usepackage{bm}
\usepackage{eucal}
\usepackage{arydshln}

\newtheorem{theorem}{Theorem}
\newtheorem{lemma}{Lemma}
\newtheorem{assumption}{Assumption}
\usepackage{pifont}
\usepackage{setspace}
\usepackage{url}
\usepackage{enumitem}
\newcommand{\circled}[1]{\textcircled{\small #1}}

\newcommand{\hetero}{heterogeneous }
\newcommand{\homo}{homogeneous }
\newcommand{\pers}{personalized }
\newcommand{\gen}{generalized }

\newcommand{\heteroN}{heterogeneity }
\newcommand{\homoN}{homogeneity }
\newcommand{\persN}{personalization }
\newcommand{\genN}{generalization }

\newcommand{\sota}{state-of-the-art }
\newcommand{\assum}{Assumption }

\newcommand{\methodname}{{\tt{pFedAFM}}}

\AtBeginDocument{%
  \providecommand\BibTeX{{%
    \normalfont B\kern-0.5em{\scshape i\kern-0.25em b}\kern-0.8em\TeX}}}

\renewcommand\footnotetextcopyrightpermission[1]{} 
\acmConference[Conference acronym 'XX]{Make sure to enter the correct
  conference title from your rights confirmation emai}{June 03--05,
  2018}{Woodstock, NY}
\begin{document}

\title[Batch-Level Heterogeneous Personalized Federated Learning]{pFedAFM: Adaptive Feature Mixture for Batch-Level Personalization in Heterogeneous Federated Learning}


\author{Liping Yi}
\email{yiliping@nbjl.nankai.edu.cn}
\orcid{0000-0001-6236-3673}
\affiliation{%
  \institution{College of Computer Science, TMCC, SysNet, DISSec, GTIISC, \\ Nankai University}
  \city{Tianjin}
  \country{China}
}

\author{Han Yu}
\email{han.yu@ntu.edu.sg}
\orcid{0000-0001-6893-8650}
\affiliation{%
  \institution{College of Computing and Data Science, Nanyang Technological University (NTU)}
  \country{Singapore}
}

\author{Chao Ren}
\email{chao.ren@ntu.edu.sg}
\orcid{0000-0001-9096-8792} 
\affiliation{%
  \institution{College of Computing and Data Science, Nanyang Technological University (NTU)}
  \country{Singapore}
}

\author{Heng Zhang}
\email{hengzhang@tju.edu.cn}
\orcid{0000-0003-4874-6162}
\affiliation{%
  \institution{College of Intelligence and Computing, Tianjin University}
  \city{Tianjin}
  \country{China}
}

\author{Gang Wang}
\email{wgzwp@nbjl.nankai.edu.cn}
\orcid{0000-0003-0387-2501}
\affiliation{%
  \institution{College of Computer Science, TMCC, SysNet, DISSec, GTIISC, \\ Nankai University}
  \city{Tianjin}
  \country{China}
}

\author{Xiaoguang Liu}
\email{liuxg@nbjl.nankai.edu.cn}
\orcid{0000-0002-9010-3278}
\affiliation{%
  \institution{College of Computer Science, TMCC, SysNet, DISSec, GTIISC, \\ Nankai University}
  \city{Tianjin}
  \country{China}
}

\author{Xiaoxiao Li}
\email{xiaoxiao.li@ece.ubc.ca}
\affiliation{%
  \institution{Department of Electrical and Computer Engineering, \\ The University of British Columbia}
  \city{Vancouver, BC}
  \country{Canada}
}

\renewcommand{\shortauthors}{L. Yi and H. Yu, et al.}

\begin{abstract}
Model-heterogeneous \pers federated learning (MHPFL) enables FL clients to train structurally different \pers models on non-independent and identically distributed (non-IID) local data.
Existing MHPFL methods focus on achieving client-level personalization, but cannot address batch-level data heterogeneity.
To bridge this important gap, we propose a model-\hetero \underline{p}ersonalized \underline{Fed}erated learning approach with \underline{A}daptive \underline{F}eature \underline{M}ixture (\methodname{}) for supervised learning tasks. It consists of three novel designs:
1) A sharing global \homo small feature extractor is assigned alongside each client's local \hetero model (consisting of a \hetero feature extractor and a prediction header) to facilitate cross-client knowledge fusion. The two feature extractors share the local \hetero model's prediction header containing rich \pers prediction knowledge to retain \pers prediction capabilities.
2) An iterative training strategy is designed to alternately train the global \homo small feature extractor and the local \hetero large model for effective global-local knowledge exchange.
3) A trainable weight vector is designed to dynamically mix the features extracted by both feature extractors to adapt to batch-level data heterogeneity.
Theoretical analysis proves that \methodname{} can converge over time. Extensive experiments on $2$ benchmark datasets demonstrate that it significantly outperforms $7$ state-of-the-art MHPFL methods, achieving up to $7.93\%$ accuracy improvement while incurring low communication and computation costs.

\end{abstract}




\maketitle

\section{Introduction}
Federated learning (FL)~\citep{1w-survey} is a distributed collaborative machine learning paradigm. It typically leverages an FL server to coordinate multiple FL clients to train a shared global model without exposing potentially sensitive local data, thereby preserving data privacy \citep{SU-Net,FedRRA,FFEDCL}.

\begin{figure}[t]
\centering
\includegraphics[width=0.8\linewidth]{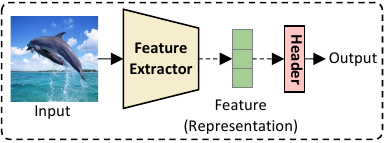}
\caption{Feature extractor and prediction header.}
\label{fig:model-split}
\end{figure}

Clients participating in FL are often devices with diverse system configurations in terms of communication bandwidth, computing power, memory and storage (\emph{a.k.a}, system heterogeneity) \citep{QSFL,FedPE}. 
They can also possess non-IID data (\emph{a.k.a}, data heterogeneity) due to differences in usage environments \citep{pFedKT,pFedLHNs}.
Classical FL aggregation approaches, such as {\tt{FedAvg}}~\citep{FedAvg}, require the server to broadcast the global model to the clients for training on local data. The trained local models are then sent back to the server for aggregation. Thus, all clients are required to train models of the same structure.
This forces all clients to train a \homo model with a size that can be supported by the most resource-constrained client device, thereby limiting global model performance and wasting the system resources of higher-end client devices. Besides, a single global model for all clients often cannot perform well on each client's non-IID local data \citep{Non-IID}. Furthermore, clients with proprietary models might be reluctant to share their model structures with others due to intellectual property protection concerns \citep{FedGH,FedSSA}. In response to these challenges, the field of Model-Heterogeneous Personalized Federated Learning (MHPFL) \citep{pFedES,FedLoRA,pFedMoE} has emerged to enable clients to train local models with diverse structures, while still joining FL.

Existing MHPFL methods take one of the following two main approaches: (1) The server or clients pruning the global model into subnets with different structures and aggregating them by parameter positions, constraining the relationship between the global model and client models \cite{FjORD,HeteroFL}; and
(2) clients training structurally unrelated local models, supporting higher levels of model \heteroN \citep{FedGH}. Our research focuses on the second approach due to its practicality. It leverages three main techniques: knowledge distillation, model mixture and mutual learning. Knowledge distillation-based MHPFL methods \cite{FedMD,FCCL} often rely on the availability of a suitable public dataset. This is hard to achieve in practice. 
Model mixture-based MHPFL methods \citep{LG-FedAvg,FedRep} share the \homo parts of \hetero local models for FL aggregation. They often face model performance bottlenecks and exposure to partial model structures (which can compromise privacy). Mutual learning-based MHPFL methods \citep{FedKD,FedAPEN} equip each client's \hetero local model with an additional small \homo model. The two models exchange local and global knowledge via mutual learning. The small \homo local models are used for FL aggregation at the server. However, model performance improvements tend to be small due to the limitations of mutual learning.


Existing MHPFL methods are designed for client-level personalization. They cannot deal with the problem of data distribution variations across local training batches (\emph{i.e.}, \textit{batch-level data heterogeneity}), which often take place.
To bridge this important gap, we propose the model-\hetero \underline{p}ersonalized \underline{Fed}erated learning approach with \underline{A}daptive \underline{F}eature \underline{M}ixture (\methodname{}) for supervised learning tasks. It consists of three novel designs: 
    \textbf{(1) Model Architecture}: We split each client's local \hetero model into two parts: 1) a \hetero feature extractor, and 2) a prediction header as in Figure~\ref{fig:model-split}. A shared global small \homo feature extractor is additionally assigned to each client for cross-client knowledge sharing. The two feature extractors share the same prediction header.
     \textbf{(2) Model Training}: We design an iterative training method to train the two models alternatively for bidirectionally transferring the global \gen and local \pers knowledge between the server and clients.
      \textbf{(3) Adaptive Feature Mixing}: For each training batch, each client trains a \pers trainable weight vector to mix the representations extracted by the two feature extractors. The weighted mixed representations are processed by the header for prediction. Then, the trainable weight vector and the local \hetero model are updated simultaneously. In this way, \methodname{} achieves batch-level \persN through fine-grained representation mixing in response to local data distribution variations across different training batches.

Theoretical analysis proves that it converges over time with a $\mathcal{O}(1/T)$ non-convex convergence rate. Experiments on $2$ benchmark datasets across $7$ state-of-the-art MHPFL methods demonstrate its superiority in model performance, achieving up to $7.93\%$ accuracy improvement with low communication and computation costs.

    
    

\section{Related Work}\label{sec:related}
MHPFL works can be classified into two families: (1) Clients training \hetero subnets of the global model: these local \hetero subnets are aggregated at the server by parameter ordinates. These methods (\emph{e.g.}, {\tt{FedRolex}} \citep{FedRolex}, {\tt{FLASH}}~\citep{FLASH}, {\tt{InCo}}~\citep{InCo}, {\tt{HeteroFL}} \citep{HeteroFL}, {\tt{FjORD}} \citep{FjORD}, {\tt{HFL}} \citep{HFL}, {\tt{Fed2}} \citep{Fed2}, {\tt{FedResCuE}} \citep{FedResCuE}) often utilize model pruning techniques to construct \hetero subnets, incurring high computational overheads and limiting the relationships between client models. (2) Clients training \hetero models with diverse structures: it offers high flexibility in model heterogeneity. Our work falls under the latter family, which can be further divided into three types based on the techniques adopted.

\textbf{MHPFL with Knowledge Distillation.} Most of these methods ({\tt{Cronus}} \citep{Cronus}, {\tt{FedGEMS}} \citep{FedGEMS}, {\tt{Fed-ET}} \citep{Fed-ET}, {\tt{FSFL}} \citep{FSFL}, {\tt{FCCL}} \citep{FCCL}, {\tt{DS-FL}} \citep{DS-FL}, {\tt{FedMD}} \citep{FedMD}, {\tt{FedKT}} \citep{FedKT}, {\tt{FedDF}} \citep{FedDF}, {\tt{FedHeNN}} \citep{FedHeNN}, {\tt{FedKEM}} \citep{FedKEM}, {\tt{KRR-KD}} \citep{KRR-KD}, {\tt{FedAUX}} \citep{FEDAUX}, {\tt{CFD}} \citep{CFD}, {\tt{pFedHR}} \citep{pFedHR}, {\tt{FedKEMF}} \citep{FedKEMF} and {\tt{KT-pFL}} \citep{KT-pFL}) rely on an additional labeled or unlabelled public data with a similar distribution to the client's local data, which is not often accessible in practice due to data privacy concerns. Communication and computational overhead involving all samples of the public dataset is high.
To avoid using an extra public dataset, some methods (\emph{e.g.}, {\tt{FedGD}} \citep{FedGD}, {\tt{FedZKT}} \citep{FedZKT}, {\tt{FedGen}} \citep{FedGen}) train a generator to generate synthetic public data, which introduces high training costs. Remaining methods (\emph{e.g.},  {\tt{HFD}} \citep{HFD1,HFD2}, {\tt{FedGKT}} \citep{FedGKT}, {\tt{FD}} \citep{FD}, {\tt{FedProto}} \citep{FedProto}, and {\tt{FedGH}} \citep{FedGH}) aggregate seen-class information from clients by class at the server, which risks privacy leakage due to uploading class information.

\textbf{MHPFL with Model Mixture.} These methods divide a client's local model as two parts: a feature extractor and a prediction header. Different clients may possess \hetero feature extractors and \homo prediction headers, with the latter being shared for aggregation (\emph{e.g.}, {\tt{FedMatch}} \citep{FedMatch}, {\tt{FedRep}} \citep{FedRep}, {\tt{FedBABU}} \citep{FedBABU} and {\tt{FedAlt/FedSim}} \citep{FedAlt/FedSim}). In contrast, other methods ({\tt{FedClassAvg}} \citep{FedClassAvg}, {\tt{LG-FedAvg}} \citep{LG-FedAvg} and {\tt{CHFL}} \citep{CHFL}) share the \homo feature extractors, while keeping \hetero prediction header private. Sharing parts of the local models leaks partial model structure and leads to performance bottlenecks.

\textbf{MHPFL with Mutual Learning.} These methods ({\tt{FML}} \citep{FML} and {\tt{FedKD}} \citep{FedKD}) add a small \homo model to each client's local \hetero model. Clients train them through mutual learning, and the server aggregates the locally trained small \homo models. However, the improperly trained shared small \homo model during early rounds of FL might negatively affect model convergence. 
To solve this issue, a recent MHPFL research - {\tt{FedAPEN}}~\citep{FedAPEN}, based on mutual learning, allows each client to first train a \pers learnable weight on partial local data for the output logits of the local \hetero model. Then, it performs $1-$ operation on this weight as the weight of the \homo small model's output logits. Fixing the two weights, it then trains these models via end-to-end mutual learning on the remaining local data. It achieves client-level \persN by training a learnable weight for the output ensemble.
However, the fixed weight during model training might negatively impact model performance due to the distribution divergences between the partial data for training the learnable weight versus the remaining data. This can be further exacerbated by the batch-level data heterogeneity issue. Moreover, different dimensions of the model output logits contain different semantic information. Sharing the same weight for all output dimensions fails to exploit the differences in the importance of semantic information across different dimensions, thereby limiting model performance.

Existing MHPFL methods are not designed to deal with the issue of batch-level data heterogeneity. To the best of our knowledge, \methodname{} is the first approach to bridge this important gap by fine-grained dimension-level representation mixing adaptive to local data distribution variations across different training batches.


\section{Problem Formulation}
Under \methodname{}, a central FL server coordinates $N$ FL clients with \hetero local models for collaborative training. In each communication round, a fraction $C$ of the $N$ clients join FL model training. The selected client set is denoted as $\boldsymbol{\mathcal{S}}$, $|\boldsymbol{\mathcal{S}}|=C \cdot N =K$. The local \hetero model $\mathcal{F}_k(\omega_k)$ held by client $k$  ($\mathcal{F}_k(\cdot)$ is the heterogeneous model structure, $\omega_k$ are the \pers model parameters) consists of a feature extractor $\mathcal{F}_k^{ex}(\omega_k^{ex})$ and a prediction header $\mathcal{F}_k^{hd}(\omega_k^{hd})$. Client $k$'s non-IID local dataset $D_k$ follows distribution $P_k$. 
To achieve knowledge sharing across \hetero clients, a small \homo global shared feature extractor $\mathcal{G}(\theta)$ ($\mathcal{G}(\cdot)$ is the \homo model structure, $\theta$ are parameters) is assigned to each client. After training the two models locally via the proposed iterative training method, $\mathcal{G}(\theta_k)$ is uploaded to the server for aggregation. Therefore, the training objective of \methodname{} is to minimize the sum of the loss of the combined complete model $\mathcal{H}_k(h_k)=\mathcal{G}(\theta)\circ\mathcal{F}_k(\omega_k)$ of all clients on local data $D_k$:
\begin{equation}\label{eq:init-def}
\min _{\theta, \omega_{0, \ldots, N-1}} \sum_{k=0}^{N-1} \ell(\mathcal{H}_k(D_k ; \theta \circ \omega_k)).
\end{equation}

\section{The Proposed Approach}
\methodname{} consists of three key design considerations as follows.

\textbf{(1) Model Architecture.} We cannot directly aggregate local models with \hetero structures at the server. Since the representations extracted by the feature extractor contain richer semantic information than model output logits and the local \hetero model's header closer to model output carries sufficient \pers prediction information, we additionally assign a global shared \homo small feature extractor to all clients, and the \homo small feature extractor and the local \hetero model's feature extractor shares the local \hetero model's prediction header. The locally updated \homo small feature extractors are aggregated at the server for cross-client knowledge fusion.

\textbf{(2) Model Training.} An intuitive way to train the \homo small feature extractor and the local \hetero model is updating them simultaneously in an end-to-end manner. However, the immature global \homo feature extractor in early rounds may harm model performance within end-to-end training, and the complete global information is only transferred to clients within the first local iteration. To address these concerns, we devise an iterative training approach. Firstly, we freeze the global shared \homo small feature extractor, and train the local \hetero model's feature extractor and prediction header, transferring knowledge from global to local. Then, we freeze the local \hetero model's prediction header and train the global \homo small feature extractor, transferring knowledge from local to global.

\textbf{(3) Adaptive Feature Mixing.} In the first step of iterative training, we train a pair of trainable weight vectors for weighting the representations extracted by the global \homo small feature extractor and the local \hetero feature extractor. Each dimension of the weight vector corresponds to a dimension of the representation, achieving a finer-grained feature mixture. The weighted mixed representations absorb the \gen feature information from the global \homo small feature extractor and the \pers feature information from the local \hetero feature extractor.
Moreover, the trainable weight vector is updated simultaneously with the local \hetero model, which dynamically balances the \genN and \persN of local models adaptive to the distribution variations across different batches, \emph{i.e.}, achieving adaptive batch-level personalization.

\begin{figure}[t]
\centering
\includegraphics[width=\linewidth]{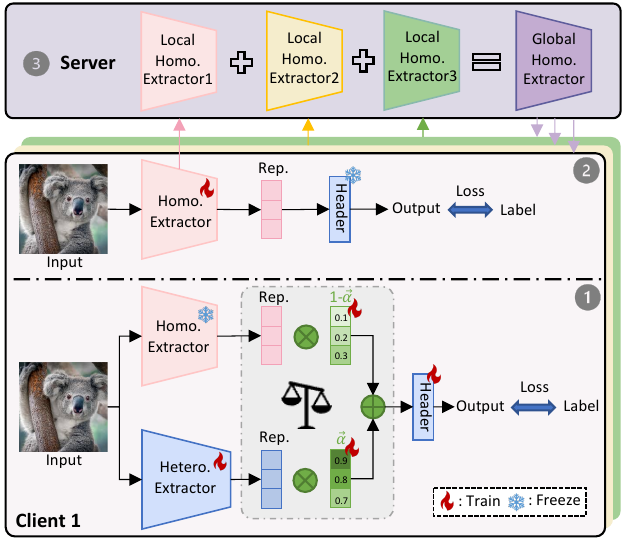}
\caption{Workflow of \methodname{}.}
\label{fig:pFedAFM}
\end{figure}

\textbf{Overview.} With the above considerations, we design \methodname{} to perform the following steps in each training round (Figure~\ref{fig:pFedAFM}).

\begin{enumerate}[label=\circled{\arabic*}]
    \item Each client receives the global \homo small feature extractor from the server and freezes its parameters to train the local \hetero model feature extractor and prediction header. In a training batch, a pair of trainable weight vectors weigh the representations extracted by the two feature extractors on local data samples of this batch to produce the mixed representations, which are further processed by the local \hetero model's perdition header to output predictions. The loss between the predictions and the labels is used to update the parameters of the local \hetero model and the trainable weight vectors.

    \item Each client freezes the updated parameters of the local \hetero model's prediction header and feeds local data samples into the global \homo feature extractor to extract representations. They are further fed into the frozen prediction header to produce predictions. The loss between the predictions and the labels is used to update the parameters of the \homo small feature extractor.
    
    \item The updated local \homo small feature extractor is uploaded to the server for aggregation to produce the new global \homo feature extractor. It is then distributed to clients participating in the next round of FL.
    
\end{enumerate}
The three steps repeat until all clients' local combined \hetero models (the mixture of the local \hetero model and the global \homo small feature extractor) converge, which is used to perform inference after FL training. The details of \methodname{} are given in Algorithm~\ref{alg:pFedAFM}.

\begin{algorithm}[t]
  \caption{\methodname{}}
  \label{alg:pFedAFM}
  \textbf{Input}: $N$, total number of clients; $K$, number of selected clients in one round; $T$, number of communication rounds; $\eta_\omega$, learning rate of local heterogeneous models; $\eta_{\boldsymbol{\alpha}}$, learning rate of trainable weight vector; $\eta_\theta$, learning rate of homogeneous feature extractor. \\
  \vspace{1em}
     Randomly initialize global homogeneous feature extractor $\mathcal{G}(\theta^\mathbf{0})$, local trainable weight vectors $[\boldsymbol{\alpha}_0,\ldots,\boldsymbol{\alpha}_{N-1}]=\boldsymbol{1}$, and local heterogeneous models {\small$[\mathcal{F}_0(\omega_0^0),\ldots,\mathcal{F}_{N-1}(\omega_{N-1}^0)]$}. \\
    \For{$t=1$ {\bfseries to} $T-1$}{
     // \textbf{Server Side}: \\
         $\boldsymbol{\mathcal{S}}^t \gets$  Randomly select $K\leqslant N$ clients to join FL; \\
         Broadcast the global homogeneous feature extractor $\theta^{t-1}$ to selected $K$ clients; \\ 
         $\theta_k^t \gets$ \textbf{ClientUpdate}$(\theta^{t-1})$; \\
         \begin{tcolorbox}[colback=ylp_color2,
                  colframe=ylp_color1,
                  width=7.3cm,
                  height=1cm,
                  arc=1mm, auto outer arc,
                  boxrule=1pt,
                  left=0pt,right=0pt,top=0pt,bottom=0pt,
                 ]
         \textbf{/* Aggregate Homogeneous Feature Extractors */} \\
        $\theta^t=\sum_{k \in \boldsymbol{\mathcal{S}}^t} \frac{n_k}{n} \theta_k^t$. \\
        \end{tcolorbox}
         // \textbf{ClientUpdate}:  \\
         Receive the global homogeneous feature extractor $\theta^{t-1}$ from the server; \\
            \For{$k \in \boldsymbol{\mathcal{S}}^t$}{
              \begin{tcolorbox}[colback=ylp_color2,
                  colframe=ylp_color1,
                  width=7.3cm,
                  height=6.8cm,
                  arc=1mm, auto outer arc,
                  boxrule=1pt,
                  left=0pt,right=0pt,top=0pt,bottom=0pt,
                 ]
             \textbf{/* Freeze Homo. Extractor, Train Hetero. Model */} \\
                  \For{$(\boldsymbol{x}_i,y_i)\in D_k$}{
                    $\boldsymbol{\mathcal{R}}_{i}^{\mathcal{G}}=\mathcal{G}(\boldsymbol{x}_i ; \theta^{t-1}), 
\boldsymbol{\mathcal{R}}_{i}^{\mathcal{F}_{k}}=\mathcal{F}_k^{e x}(\boldsymbol{x}_i ; \omega_k^{e x, t-1})$;\\
                    $\boldsymbol{\mathcal{R}}_{i}= \boldsymbol{\mathcal{R}}_{i}^{\mathcal{G}} \cdot (\boldsymbol{1}-\boldsymbol{\alpha}_{k,i}^{t-1}) + \boldsymbol{\mathcal{R}}_{i}^{\mathcal{F}_{k}} \cdot \boldsymbol{\alpha}_{k,i}^{t-1}$;\\
                    $\hat{y}_i=\mathcal{F}_k^{h d}(\boldsymbol{\mathcal{R}}_{i} ; \omega_k^{h d, t-1})$; \\
                    $\ell_i=\ell(\hat{y}_i, y_i)$; \\
                    $\omega_k^t \gets \omega_k^{t-1}-\eta_\omega \nabla \ell_i$; \\
                    $\boldsymbol{\alpha}_{k,i}^t \gets \boldsymbol{\alpha}_{k,i}^{t-1}-\eta_{\boldsymbol{\alpha}} \nabla \ell_i$;  
                    }
             \textbf{/* Freeze Hetero. Model, Train Homo. Extractor */} \\
                \For{$(\boldsymbol{x}_i,y_i)\in D_k$}{
                    $\boldsymbol{\mathcal{R}}_{i}=\mathcal{G}(\boldsymbol{x}_i ; \theta^{t-1})$; \\
                    $\hat{y}_i=\mathcal{F}_k^{h d}(\boldsymbol{\mathcal{R}}_{i} ; \omega_k^{h d, t})$; \\
                    $\ell_i=\ell(\hat{y}_i, y_i)$; \\
                    $\theta^t \gets \theta^{t-1}-\eta_\theta \nabla \ell_i$;
                  }
                \end{tcolorbox}
                Upload trained local homogeneous feature extractor $\theta_k^t$ to the server.
            }
    }
     \textbf{Output:} heterogeneous local complete mixed models 
     \mbox{
      \begin{tcolorbox}[colback=ylp_color2,
                  colframe=ylp_color1,
                  width=8cm,
                  height=0.55cm,
                  arc=1mm, auto outer arc,
                  boxrule=1pt,
                  left=0pt,right=0pt,top=0pt,bottom=0pt,
                 ]
     {\footnotesize$[(\mathcal{G}(\theta^{T-1})\circ\mathcal{F}_0(\omega_0^{T-1})|\boldsymbol{\alpha}_0^{T-1}),\ldots,(\mathcal{G}(\theta^{T-1})\circ\mathcal{F}_{N-1}(\omega_{N-1}^{T-1})|\boldsymbol{\alpha}_{N-1}^{T-1})]$.}
      \end{tcolorbox}
      }
\end{algorithm}

\subsection{Knowledge Exchange by Iterative Training}

\subsubsection{Local Heterogeneous Model Training with Adaptive Batch-Level Personalization}
In the $t$-th communication round, for global-to-local knowledge transfer, a client $k$ freezes the received global \homo small feature extractor $\mathcal{G}(\theta^{{t}-\mathbf{1}})$, and trains its local \hetero model's feature extractor $\mathcal{F}_k^{ex}(\omega_k^{ex,t-1})$ and prediction header $\mathcal{F}_k^{hd}(\omega_k^{hd,t-1})$.
Specifically, for each data sample $(\boldsymbol{x}_i,y_i)\in D_k$, $\boldsymbol{x}_i$ is fed into the frozen $\mathcal{G}(\theta^{{t}-\mathbf{1}})$, which contains \gen feature information across all labels, to extract \gen representations. It is also fed into $\mathcal{F}_k^{ex}(\omega_k^{ex,t-1})$, which contains local \pers feature information across seen classes, to extract \pers representations.
\begin{equation}
\boldsymbol{\mathcal{R}}_{i}^{\mathcal{G}}=\mathcal{G}(\boldsymbol{x}_i ; \theta^{t-1}), 
\boldsymbol{\mathcal{R}}_{i}^{\mathcal{F}_{k}}=\mathcal{F}_k^{e x}(\boldsymbol{x}_i ; \omega_k^{e x, t-1}).
\end{equation}

To effectively balance the \genN and \persN of local models, we learn a trainable weight vector $\boldsymbol{\alpha}_{k,i}^{t-1}$ to weigh the representations $\boldsymbol{\mathcal{R}}_{i}^{\mathcal{F}_{k}}$ extracted by the local \hetero feature extractor. $\boldsymbol{1}-\boldsymbol{\alpha}_{k,i}^{t-1}$ is the weight vector for weighing the representations $\boldsymbol{\mathcal{R}}_{i}^{\mathcal{G}}$ extracted by the frozen global \homo small feature extractor. The two weight vectors have the same dimensions as the representations. The representations extracted by two feature extractors are dot-multiplied with their corresponding weight vectors, then summed element-wise to produce the mixed representations:
\begin{equation}
\boldsymbol{\mathcal{R}}_{i}= \boldsymbol{\mathcal{R}}_{i}^{\mathcal{G}} \cdot (\boldsymbol{1}-\boldsymbol{\alpha}_{k,i}^{t-1}) + \boldsymbol{\mathcal{R}}_{i}^{\mathcal{F}_{k}} \cdot \boldsymbol{\alpha}_{k,i}^{t-1}.
\end{equation}
Mixing the representations extracted by two feature extractors via weighted summation requires the dimension of the last linear layer of the two feature extractors to be the same. For a client with a structure-agnostic local model (\emph{i.e.}, black box), if its penultimate layer is linear (\emph{i.e.}, the last linear layer is the prediction header), it needs to modify the penultimate layer dimension to be the same as the last linear layer dimension of the global \homo feature extractor. If its penultimate layer is non-linear, it needs to add a linear layer with an identical dimension to the global \homo feature extractor's last linear layer ahead of the prediction header. This dimension-matching step is straightforward to implement.

The mixed representation $\boldsymbol{\mathcal{R}}_{i}$ fuses global \gen and local \pers feature information. The mixed representation $\boldsymbol{\mathcal{R}}_{i}$ is then fed into the local \hetero model's prediction header $\mathcal{F}_k^{hd}(\omega_k^{hd,t-1})$ to produce predictions:
\begin{equation}
\hat{y}_i=\mathcal{F}_k^{h d}(\boldsymbol{\mathcal{R}}_{i} ; \omega_k^{h d, t-1}).
\end{equation}
Then, the loss (\emph{e.g.}, cross-entropy loss \citep{CEloss} in classification tasks) between prediction $\hat{y}_i$ and label $y_i$ is computed as:
\begin{equation}
 \ell_i=\ell(\hat{y}_i, y_i).   
\end{equation}
The loss is used to update the parameters of the local \hetero model and the pair of trainable weight vectors via gradient descent (\emph{e.g.}, SGD optimizer~\citep{SGD}):
\begin{equation}
\begin{aligned}
\omega_k^t &\gets \omega_k^{t-1}-\eta_\omega \nabla \ell_i, \\
\boldsymbol{\alpha}_{k,i}^t &\gets \boldsymbol{\alpha}_{k,i}^{t-1}-\eta_{\boldsymbol{\alpha}} \nabla \ell_i,   
\end{aligned}
\end{equation}
where $\eta_\omega$ and $\eta_{\boldsymbol{\alpha}}$ are the learning rates of the local \hetero model and the trainable weight vectors. 
Therefore, the trainable weight vectors dynamically balance the \genN and \persN as the data distribution varies across different training batches, thereby achieving adaptive batch-level personalization.

\subsubsection{Global Homogeneous Feature Extractor Training}
After training the local \hetero model, a client $k$ freezes its local header $\mathcal{F}_k^{hd}(\omega_k^{hd,t})$ and trains the global \homo small feature extractor on local data $D_k$ for local-to-global knowledge transfer.

Specifically, client $k$ inputs each sample $(\boldsymbol{x}_i,y_i)\in D_k$ into the global \homo feature extractor to extract its representation:
\begin{equation}
\boldsymbol{\mathcal{R}}_{i}=\mathcal{G}(\boldsymbol{x}_i ; \theta^{t-1}).
\end{equation}
The representation is then fed into the frozen local \hetero model's prediction header $\mathcal{F}_k^{hd}(\omega_k^{hd,t})$ to produce a prediction:
\begin{equation}
\hat{y}_i=\mathcal{F}_k^{h d}(\boldsymbol{\mathcal{R}}_{i} ; \omega_k^{h d, t}).
\end{equation}
The loss between the prediction and the label is then computed as:
\begin{equation}
 \ell_i=\ell(\hat{y}_i, y_i).
\end{equation}
The \homo feature extractor is updated via gradient descent:
\begin{equation}
\theta^t \gets \theta^{t-1}-\eta_\theta \nabla \ell_i,
\end{equation}
where $\eta_\theta$ is the learning rate of the \homo feature extractor. We set $\eta_\theta = \eta_\omega$ to ensure stable convergence of local models.

\subsection{Knowledge Fusion across Clients}
Client $k$ sends its updated local \homo extractor $\theta_k^t$ to the server for aggregation to fuse cross-client knowledge:
\begin{equation}
\theta^t=\sum_{k \in \boldsymbol{\mathcal{S}^t}} \frac{n_k}{n} \theta_k^t,
\end{equation}
where $n_k=|D_k|$ is the size of client $k$'s local dataset $D_k$, and $n$ is the total size of all clients' local datasets.

\noindent\textbf{Summary.} We view local \homo small feature extractors as the carriers of local knowledge and enable knowledge fusion across \hetero local models by sharing them. Through iterative training of the global shared small \homo feature extractor and local \hetero models, we achieve effective bidirectional knowledge transfer. Through dimension-wise mixing of \gen features and \pers features, \methodname{} adaptively balances model \genN and \persN at the batch level. 
Hence, the objective of \methodname{} in Eq.~(\ref{eq:init-def}) can be refined as:
\begin{equation}\label{eq:final-def}
\min _{\theta, \omega_{0, \ldots, N-1}} \sum_{k=0}^{N-1} \ell(\mathcal{H}_k(D_k ; \theta \cdot (\boldsymbol{1}-\boldsymbol{\alpha}_{k}) + \omega_k \cdot \boldsymbol{\alpha}_{k})).
\end{equation}


\section{Convergence Analysis}
We clarify some notations.
$t \in \{0,\ldots,T-1\}$ is the $t$-th communication round. 
$e\in\{0,1,\ldots,E\}$ is the $e$-th local iteration.
$tE+0$ denotes the start of the $(t+1)$-th round in which client $k$ in the $(t+1)$-th round receives the small \homo feature extractor $\mathcal{G}(\theta^{t})$ from the server.
$tE+e$ is the $e$-th local iteration in the $(t+1)$-th round.
$tE+E$ is the last local iteration in the $(t+1)$-th round. After that, client $k$ sends its local updated small \homo feature extractor the server for aggregation.
$\mathcal{H}_{k}(h_k)$ is client $k$'s entire local model consisting of the global small \homo feature extractor $\mathcal{G}(\theta)$ and the local \hetero model $\mathcal{F}_{k}(\omega_k)$ weighed by the trainable weight vector $\boldsymbol{\alpha}_k$, \emph{i.e.}, $\mathcal{H}_{k}(h_k) = (\mathcal{G}(\theta) \circ \mathcal{F}_{k}(\omega_k) | \boldsymbol{\alpha}_k)$.
$\eta$ is the learning rate of client $k$'s local model $\mathcal{H}_{k}(h_k)$, consisting of $\{\eta_\theta,\eta_\omega,\eta_{\boldsymbol{\alpha}}\}$.

\begin{assumption}\label{assump:Lipschitz}
\textbf{Lipschitz Smoothness}. The gradients of client $k$'s entire local heterogeneous model $h_k$ are $L1$--Lipschitz smooth \cite{FedProto},
\begin{equation}\label{eq:smmoth}
\footnotesize
\begin{gathered}
\|\nabla \mathcal{L}_k^{t_1}(h_k^{t_1} ; \boldsymbol{x}, y)-\nabla \mathcal{L}_k^{t_2}(h_k^{t_2} ; \boldsymbol{x}, y)\| \leqslant L_1\|h_k^{t_1}-h_k^{t_2}\|, \\
\forall t_1, t_2>0, k \in\{0,1, \ldots, N-1\},(\boldsymbol{x}, y) \in D_k.
\end{gathered}
\end{equation}
The above formulation can be re-expressed as:
\begin{equation}
\footnotesize
\mathcal{L}_k^{t_1}-\mathcal{L}_k^{t_2} \leqslant\langle\nabla \mathcal{L}_k^{t_2},(h_k^{t_1}-h_k^{t_2})\rangle+\frac{L_1}{2}\|h_k^{t_1}-h_k^{t_2}\|_2^2 .
\end{equation}
\end{assumption}

\begin{assumption} \label{assump:Unbiased}
\textbf{Unbiased Gradient and Bounded Variance}. Client $k$'s random gradient $g_{h,k}^t=\nabla \mathcal{L}_k^t(h_k^t; \mathcal{B}_k^t)$ ($\mathcal{B}$ is a batch of local data) is unbiased, 
\begin{equation}
\footnotesize
\mathbb{E}_{\mathcal{B}_k^t \subseteq D_k}[g_{h,k}^t]=\nabla \mathcal{L}_k^t(h_k^t),
\end{equation}
and the variance of random gradient $g_{h,k}^t$ is bounded by:
\begin{equation}
\footnotesize
\begin{split}
\mathbb{E}_{\mathcal{B}_k^t \subseteq D_k}[\|\nabla \mathcal{L}_k^t(h_k^t ; \mathcal{B}_k^t)-\nabla \mathcal{L}_k^t(h_k^t)\|_2^2] \leqslant \sigma^2.
\end{split}
\end{equation}    
\end{assumption}

\begin{assumption} \label{assump:BoundedVariation}
\textbf{Bounded Parameter Variation}. The parameter variations of the small \homo feature extractor $\theta_k^t$ and $\theta^t$ before and after aggregation at the FL server is bounded by:
\begin{equation}
\footnotesize
     {\|\theta^t - \theta_k^{t}\|}_2^2 \leq \delta^2.
\end{equation}
\end{assumption} 

Based on the above assumptions, we can derive the following Lemma and Theorem. Detailed proofs are given in Appendix~\ref{app:proof}.

\begin{lemma}\label{lemma:localtraining}
    \textbf{Local Training.} Given Assumptions~\ref{assump:Lipschitz} and \ref{assump:Unbiased}, the loss of an arbitrary client's local model $h$ in the $(t+1)$-th local training round is bounded by:
    \begin{equation}
    \footnotesize
        \mathbb{E}[\mathcal{L}_{(t+1) E}] \leq \mathcal{L}_{t E+0}+(\frac{L_1 \eta^2}{2}-\eta) \sum_{e=0}^E\|\nabla \mathcal{L}_{t E+e}\|_2^2+\frac{L_1 E \eta^2 \sigma^2}{2}. 
    \end{equation}
\end{lemma}

\begin{lemma}\label{lemma:aggregation}
\textbf{Model Aggregation.} Given Assumptions \ref{assump:Unbiased} and \ref{assump:BoundedVariation}, after the $(t+1)$-th local training round, the loss of any client before and after aggregating the small \homo feature extractors at the FL server is bounded by:
\begin{equation}
 \footnotesize   \mathbb{E}[\mathcal{L}_{(t+1)E+0}]\le\mathbb{E}[\mathcal{L}_{tE+1}]+{\eta\delta}^2.
\end{equation}
\end{lemma}

\begin{theorem}\label{theorem:one-round}
\textbf{One Complete Round of FL}. Based on Lemma~\ref{lemma:localtraining} and Lemma~\ref{lemma:aggregation}, for any client, after local training, model aggregation and receiving the new global \homo feature extractor, we have:
\begin{equation}
\footnotesize
\mathbb{E}[\mathcal{L}_{(t+1) E+0}] \leq \mathcal{L}_{t E+0}+(\frac{L_1 \eta^2}{2}-\eta) \sum_{e=0}^E\|\nabla \mathcal{L}_{t E+e}\|_2^2+\frac{L_1 E \eta^2 \sigma^2}{2}+\eta \delta^2.
\end{equation}
\end{theorem}

\begin{theorem}\label{theorem:non-convex}
\textbf{Non-convex Convergence Rate of pFedAFM.} With Theorem~\ref{theorem:one-round}, for any client and an arbitrary constant $\epsilon>0$, the following holds:
\begin{equation}
\footnotesize
\begin{aligned}
\frac{1}{T} \sum_{t=0}^{T-1} \sum_{e=0}^{E-1}\|\nabla \mathcal{L}_{t E+e}\|_2^2 &\leq \frac{\frac{1}{T} \sum_{t=0}^{T-1}[\mathcal{L}_{t E+0}-\mathbb{E}[\mathcal{L}_{(t+1) E+0}]]+\frac{L_1 E \eta^2 \sigma^2}{2}+\eta \delta^2}{\eta-\frac{L_1 \eta^2}{2}}<\epsilon, \\
s.t. \   &\eta<\frac{2(\epsilon-\delta^2)}{L_1(\epsilon+E \sigma^2)} .
\end{aligned}
\end{equation}
\end{theorem}
Therefore, we conclude that any client's local model can converge at a non-convex rate of $\epsilon \sim \mathcal{O}(1/T)$ in \methodname{} if the learning rates of the \homo feature extractor, local \hetero model and the trainable weight vector satisfy the above condition.

\section{Experimental Evaluation}
We implement \methodname{} and $7$ state-of-the-art MHPFL baselines with Pytorch, and evaluate performances over $2$ benchmark datasets on $4$ NVIDIA GeForce RTX $3090$ GPUs with $24$ GB memory.

\subsection{Experiment Setup}

\textbf{Datasets.} We test the performances of \methodname{} and baselines on benchmark image datasets CIFAR-10 and CIFAR-100 \footnote{\url{https://www.cs.toronto.edu/\%7Ekriz/cifar.html}} \cite{cifar} which are commonly used by FL image classification tasks. CIFAR-10 contains $6,000$ $10$-class colourful images with $32\times32$ size, with $5,000$ images for training and $1,000$ images for testing. CIFAR-100 contains $100$ classes of colourful images with $32\times32$ size. Each class contains $500$ training images and $100$ testing images.
We use two typical approaches to construct non-IID datasets. (1) Pathological \cite{pFedHN}: we assign $2$ classes of CIFAR-10 samples to each client, denoted as (non-IID: 2/10); and assign $10$ classes of CIFAR-100 samples to each client, marked as (non-IID: 10/100). 
(2) Practical \cite{FedAPEN}: we assign all classes of CIFAR-10 or CIFAR-100 samples to each client and use a Dirichlet($\gamma$) function to produce different counts of each class on different clients. A smaller $\gamma$ controls a higher non-IID degree. For each client's assigned non-IID local dataset, we further divide it into a training set and a testing set with a ratio of $8:2$ (\emph{i.e.}, they follow the same distribution).

\textbf{Base Models.} Model \homoN is a special case of model heterogeneity. We test algorithms under both model-\homo and model-\hetero FL scenarios. In the model-\homo FL scenario, we assign the same CNN-1 (Table~\ref{tab:model-structures}) to all clients. In model-\hetero FL scenario, we allocate $5$ \hetero models \{CNN-1,$\ldots$,CNN-5\} (Table~\ref{tab:model-structures}) to clients, and each client's \hetero model id is determined by client id $\% 5$. 
For the added \homo small model or \homo feature extractor in mutual learning-based {\tt{FML}}, {\tt{FedKD}}, {\tt{FedAPEN}} and \methodname{}, we choose the smallest CNN-5 or CNN-5 without the last linear layer.

\textbf{Comparison Baselines.} We compare \methodname{} with the \sota MHPFL methods belonging to the three most related heterogeneity-flexible MHPFL works outlined in Section~\ref{sec:related}.

\begin{itemize}
    \item {\tt{Standalone.}} Each client trains its local \hetero model only with local data (\emph{i.e.}, no FL).
    \item \textbf{MHPFL via Public Data-Independent Knowledge Distillation:} {\tt{FD}}~\citep{FD} and {\tt{FedProto}}~\citep{FedProto}.
    \item \textbf{MHPFL via Model Mixture:} {\tt{LG-FedAvg}}~\citep{LG-FedAvg}.
    \item \textbf{MHPFL via Mutual Learning:} {\tt{FML}}~\citep{FML}, {\tt{FedKD}}~\citep{FedKD}, and {\tt{FedAPEN}}~\citep{FedAPEN}. Our \methodname{} also belongs to this category.
\end{itemize}

\textbf{Evaluation Metrics.} We compare the performance of MHPFL methods with the following evaluation metrics:
\begin{itemize}
    \item \textbf{Model Accuracy.} We trace each client's local model's individual accuracy varied as rounds and compute each round's mean accuracy of all participating clients. We report the highest mean accuracy among all rounds and the individual accuracy of all clients in the last round.
    
    \item \textbf{Communication Cost.} We calculate the average number of parameters communicated between clients and the server in one round and monitor the rounds required to attain the specified target mean accuracy. The product of the two is the overall communication cost.
    
    \item \textbf{Computation Overhead.} We calculate the average computational FLOPs of clients in one round and monitor the rounds required for target mean accuracy. The product of the two is the overall computational overhead.
\end{itemize}

\textbf{Training Strategy.}
We use grid-search to find the optimal FL hyperparameters and specific hyperparameters for all algorithms.
For FL hyperparameters, we test them with batch size = $\{64, 128, 256, 512\}$, local epochs = $\{1, 10\}$, total rounds $T=\{100, 500\}$, SGD optimizer with a $0.01$ learning rate.
In \methodname{}, we vary $\eta_{\boldsymbol{\alpha}}=\{0.001, 0.01,0.1, 1\}$. We report the highest mean accuracy for all algorithms.

\begin{table}[t]
\centering
\caption{Structures of $5$ heterogeneous CNN models with $5 \times 5$ kernel size and $16$ or $32$ filters in convolutional layers.}
\resizebox{0.95\linewidth}{!}{%
\begin{tabular}{|l|c|c|c|c|c|}
\hline
Layer Name         & CNN-1    & CNN-2   & CNN-3   & CNN-4   & CNN-5   \\ \hline
Conv1              & 5$\times$5, 16   & 5$\times$5, 16  & 5$\times$5, 16  & 5$\times$5, 16  & 5$\times$5, 16  \\
Maxpool1              & 2$\times$2   & 2$\times$2  & 2$\times$2  & 2$\times$2  & 2$\times$2  \\
Conv2              & 5$\times$5, 32   & 5$\times$5, 16  & 5$\times$5, 32  & 5$\times$5, 32  & 5$\times$5, 32  \\
Maxpool2              & 2$\times$2   & 2$\times$2  & 2$\times$2  & 2$\times$2  & 2$\times$2  \\
FC1                & 2000     & 2000    & 1000    & 800     & 500     \\
FC2                & 500      & 500     & 500     & 500     & 500     \\
FC3                & 10/100   & 10/100  & 10/100  & 10/100  & 10/100  \\ \hline
model size & 10.00 MB & 6.92 MB & 5.04 MB & 3.81 MB & 2.55 MB \\ \hline
\end{tabular}%
}
\label{tab:model-structures}
\end{table}

\subsection{Results and Discussion}
To test the stability of MHPFL algorithms, we compare them under different client numbers $N$ and client participation rates $C$: $\{(N=10, C=100\%), (N=50, C=20\%), (N=100, C=10\%)\}$.

\subsubsection{Model-Homogeneous FL}
Table~\ref{tab:compare-homo} shows that \methodname{} reaches the highest mean accuracy across all FL settings. It improves up to $7.76\%$ accuracy than each FL setting's \sota baseline and increases up to $11.70\%$ accuracy than each FL setting's same-category (mutual learning) best baseline. This indicates that the batch-level \persN by dynamically mixing \gen and \pers features adaptive to local data distribution in \methodname{} takes a better balance between model \genN and personalization, hence promoting significant accuracy improvements.

\begin{table}[t]
\centering
\caption{Mean accuracy (\%) in model-\homo FL.}
\resizebox{\linewidth}{!}{%
\begin{tabular}{|l|cc|cc|cc|}
\hline
FL Setting                    & \multicolumn{2}{c|}{N=10, C=100\%}                                                                                                                   & \multicolumn{2}{c|}{N=50, C=20\%}                                                                                                                    & \multicolumn{2}{c|}{N=100, C=10\%}                                                                                                                   \\ \hline
Method                        & \multicolumn{1}{c|}{CIFAR-10}                                                      & CIFAR-100                                                     & \multicolumn{1}{c|}{CIFAR-10}                                                      & CIFAR-100                                                     & \multicolumn{1}{c|}{CIFAR-10}                                                      & CIFAR-100                                                     \\ \hline
Standalone                    & \multicolumn{1}{c|}{96.35}                                                         & \cellcolor[HTML]{C0C0C0}\textbf{74.32}                        & \multicolumn{1}{c|}{\cellcolor[HTML]{C0C0C0}\textbf{95.25}}                        & 62.38                                                         & \multicolumn{1}{c|}{\cellcolor[HTML]{C0C0C0}\textbf{92.58}}                        & \cellcolor[HTML]{C0C0C0}\textbf{54.93}                        \\
LG-FedAvg~\citep{LG-FedAvg}                     & \multicolumn{1}{c|}{\cellcolor[HTML]{C0C0C0}\textbf{96.47}}                        & 73.43                                                         & \multicolumn{1}{c|}{94.20}                                                         & 61.77                                                         & \multicolumn{1}{c|}{90.25}                                                         & 46.64                                                         \\
FD~\citep{FD}                            & \multicolumn{1}{c|}{96.30}                                                         & -                                                             & \multicolumn{1}{c|}{-}                                                             & -                                                             & \multicolumn{1}{c|}{-}                                                             & -                                                             \\
FedProto~\citep{FedProto}                      & \multicolumn{1}{c|}{95.83}                                                         & 72.79                                                         & \multicolumn{1}{c|}{95.10}                                                         & \cellcolor[HTML]{C0C0C0}\textbf{62.55}                        & \multicolumn{1}{c|}{91.19}                                                         & 54.01                                                         \\ \hline
FML~\citep{FML}                           & \multicolumn{1}{c|}{94.83}                                                         & 70.02                                                         & \multicolumn{1}{c|}{93.18}                                                         & 57.56                                                         & \multicolumn{1}{c|}{87.93}                                                         & 46.20                                                         \\
FedKD~\citep{FedKD}                         & \multicolumn{1}{c|}{94.77}                                                         & 70.04                                                         & \multicolumn{1}{c|}{92.93}                                                         & 57.56                                                         & \multicolumn{1}{c|}{\cellcolor[HTML]{EFEFEF}\textbf{90.23}}                        & \cellcolor[HTML]{EFEFEF}\textbf{50.99}                        \\
FedAPEN~\citep{FedAPEN}                       & \multicolumn{1}{c|}{\cellcolor[HTML]{EFEFEF}\textbf{95.38}}                        & \cellcolor[HTML]{EFEFEF}\textbf{71.48}                        & \multicolumn{1}{c|}{\cellcolor[HTML]{EFEFEF}\textbf{93.31}}                        & \cellcolor[HTML]{EFEFEF}\textbf{57.62}                        & \multicolumn{1}{c|}{87.97}                                                         & 46.85                                                         \\ \hline
\textbf{pFedAFM}               & \multicolumn{1}{c|}{\cellcolor[HTML]{9B9B9B}{\textbf{96.81}}} & \cellcolor[HTML]{9B9B9B}{\textbf{77.70}} & \multicolumn{1}{c|}{\cellcolor[HTML]{9B9B9B}{\textbf{96.58}}} & \cellcolor[HTML]{9B9B9B}{\textbf{67.63}} & \multicolumn{1}{c|}{\cellcolor[HTML]{9B9B9B}{\textbf{95.67}}} & \cellcolor[HTML]{9B9B9B}{\textbf{62.69}} \\ \hline
\textit{pFedAFM-Best B.}       & \multicolumn{1}{c|}{{\textit{0.34}}}  & { \textit{3.38}}                          & \multicolumn{1}{c|}{{\textit{1.33}}}  & {\textit{5.08}}                          & \multicolumn{1}{c|}{{\textit{3.09}}}  & {\ul \textit{7.76}}                          \\
\textit{pFedAFM-Best S.C.B.} & \multicolumn{1}{c|}{{\textit{1.43}}}  & {\textit{6.22}}                          & \multicolumn{1}{c|}{{\textit{3.27}}}  & {\textit{10.01}}                          & \multicolumn{1}{c|}{{\textit{5.44}}}  & {\ul \textit{11.70}}                          \\ \hline
\end{tabular}
} \\      
\raggedright 
\footnotesize Note: ``-'' denotes failure to converge. ``\mbox{
      \begin{tcolorbox}[colback=table_color2,
                  colframe=table_edge,
                  width=0.3cm,
                  height=0.25cm,
                  arc=0.25mm, auto outer arc,
                  boxrule=0.5pt,
                  left=0pt,right=0pt,top=0pt,bottom=0pt,
                 ]
      \end{tcolorbox}
      }'': the best MHPFL method.
``\mbox{
      \begin{tcolorbox}[colback=table_color3,
                  colframe=table_edge,
                  width=0.3cm,
                  height=0.25cm,
                  arc=0.25mm, auto outer arc,
                  boxrule=0.5pt,
                  left=0pt,right=0pt,top=0pt,bottom=0pt,
                 ]
      \end{tcolorbox}
      } Best B.'' indicates the best baseline. ``\mbox{
      \begin{tcolorbox}[colback=table_color1,
                  colframe=table_edge,
                  width=0.3cm,
                  height=0.25cm,
                  arc=0.25mm, auto outer arc,
                  boxrule=0.5pt,
                  left=0pt,right=0pt,top=0pt,bottom=0pt,
                 ]
      \end{tcolorbox}
      } Best S.C.B.'' means the best same-category baseline. 
\label{tab:compare-homo}
\end{table}

\begin{table}[t]
\centering
\caption{Mean accuracy (\%) in model-\hetero FL.}
\resizebox{\linewidth}{!}{%
\begin{tabular}{|l|cc|cc|cc|}
\hline
FL Setting                    & \multicolumn{2}{c|}{N=10, C=100\%}                                                                                                                   & \multicolumn{2}{c|}{N=50, C=20\%}                                                                                                                    & \multicolumn{2}{c|}{N=100, C=10\%}                                                                                                                   \\ \hline
Method                        & \multicolumn{1}{c|}{CIFAR-10}                                                      & CIFAR-100                                                     & \multicolumn{1}{c|}{CIFAR-10}                                                      & CIFAR-100                                                     & \multicolumn{1}{c|}{CIFAR-10}                                                      & CIFAR-100                                                     \\ \hline
Standalone                    & \multicolumn{1}{c|}{\cellcolor[HTML]{C0C0C0}{\textbf{96.53}}} & {72.53}                                  & \multicolumn{1}{c|}{{95.14}}                                  & \cellcolor[HTML]{C0C0C0}{\textbf{62.71}} & \multicolumn{1}{c|}{{91.97}}                                  & {53.04}                                  \\
LG-FedAvg~\citep{LG-FedAvg}                     & \multicolumn{1}{c|}{{96.30}}                                  & {72.20}                                  & \multicolumn{1}{c|}{{94.83}}                                  & {60.95}                                  & \multicolumn{1}{c|}{{91.27}}                                  & {45.83}                                  \\
FD~\citep{FD}                            & \multicolumn{1}{c|}{{96.21}}                                  & {-}                                      & \multicolumn{1}{c|}{{-}}                                      & {-}                                      & \multicolumn{1}{c|}{{-}}                                      & {-}                                      \\
FedProto~\citep{FedProto}                      & \multicolumn{1}{c|}{{96.51}}                                  & \cellcolor[HTML]{C0C0C0}{\textbf{72.59}} & \multicolumn{1}{c|}{\cellcolor[HTML]{C0C0C0}{\textbf{95.48}}} & {62.69}                                  & \multicolumn{1}{c|}{\cellcolor[HTML]{C0C0C0}{\textbf{92.49}}} & \cellcolor[HTML]{C0C0C0}{\textbf{53.67}} \\ \hline
FML~\citep{FML}                           & \multicolumn{1}{c|}{{30.48}}                                  & {16.84}                                  & \multicolumn{1}{c|}{{-}}                                      & {21.96}                                  & \multicolumn{1}{c|}{{-}}                                      & {15.21}                                  \\
FedKD~\citep{FedKD}                         & \multicolumn{1}{c|}{\cellcolor[HTML]{EFEFEF}{\textbf{80.20}}} & \cellcolor[HTML]{EFEFEF}{\textbf{53.23}} & \multicolumn{1}{c|}{\cellcolor[HTML]{EFEFEF}{\textbf{77.37}}} & \cellcolor[HTML]{EFEFEF}{\textbf{44.27}} & \multicolumn{1}{c|}{\cellcolor[HTML]{EFEFEF}{\textbf{73.21}}} & \cellcolor[HTML]{EFEFEF}{\textbf{37.21}} \\
FedAPEN~\citep{FedAPEN}                       & \multicolumn{1}{c|}{{\textbf{-}}}                             & {\textbf{-}}                             & \multicolumn{1}{c|}{{\textbf{-}}}                             & {\textbf{-}}                             & \multicolumn{1}{c|}{{-}}                                      & {-}                                      \\ \hline
\textbf{\methodname{}}               & \multicolumn{1}{c|}{\cellcolor[HTML]{9B9B9B}{\textbf{96.76}}} & \cellcolor[HTML]{9B9B9B}{\textbf{77.30}} & \multicolumn{1}{c|}{\cellcolor[HTML]{9B9B9B}{\textbf{96.69}}} & \cellcolor[HTML]{9B9B9B}{\textbf{67.95}} & \multicolumn{1}{c|}{\cellcolor[HTML]{9B9B9B}{\textbf{95.68}}} & \cellcolor[HTML]{9B9B9B}{\textbf{61.60}} \\ \hline
\textit{\methodname{}-Best B.}       & \multicolumn{1}{c|}{{\textit{0.23}}}  & {{\textit{4.71}}}                    & \multicolumn{1}{c|}{{\textit{1.21}}}  & {\textit{5.24}}                          & \multicolumn{1}{c|}{{\textit{3.19}}}  & {\ul \textit{7.93}}                          \\
\textit{\methodname{}-Best S.C.B.} & \multicolumn{1}{c|}{{\textit{16.56}}} & {{\textit{24.07}}}                   & \multicolumn{1}{c|}{{\textit{19.32}}} & {\textit{23.68}}                         & \multicolumn{1}{c|}{{\textit{22.47}}} & {\ul \textit{24.39}}                         \\ \hline
\end{tabular}
} \\
\raggedright 
\footnotesize Note: ``-'' denotes failure to converge. ``\mbox{
      \begin{tcolorbox}[colback=table_color2,
                  colframe=table_edge,
                  width=0.3cm,
                  height=0.25cm,
                  arc=0.25mm, auto outer arc,
                  boxrule=0.5pt,
                  left=0pt,right=0pt,top=0pt,bottom=0pt,
                 ]
      \end{tcolorbox}
      }'': the best MHPFL method.
``\mbox{
      \begin{tcolorbox}[colback=table_color3,
                  colframe=table_edge,
                  width=0.3cm,
                  height=0.25cm,
                  arc=0.25mm, auto outer arc,
                  boxrule=0.5pt,
                  left=0pt,right=0pt,top=0pt,bottom=0pt,
                 ]
      \end{tcolorbox}
      } Best B.'' indicates the best baseline. ``\mbox{
      \begin{tcolorbox}[colback=table_color1,
                  colframe=table_edge,
                  width=0.3cm,
                  height=0.25cm,
                  arc=0.25mm, auto outer arc,
                  boxrule=0.5pt,
                  left=0pt,right=0pt,top=0pt,bottom=0pt,
                 ]
      \end{tcolorbox}
      } Best S.C.B.'' means the best same-category baseline. 

\label{tab:compare-hetero}
\end{table}

\subsubsection{Model-Heterogeneous FL}
We compare \methodname{} with baselines under model-\hetero FL scenarios as follows. 

\begin{figure}[t]
\centering
\begin{minipage}[t]{0.33\linewidth}
\centering
\includegraphics[width=1.2in]{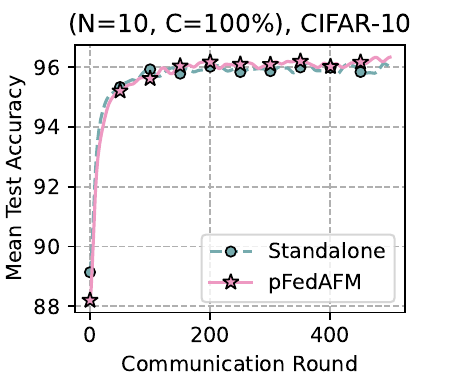}
\end{minipage}%
\begin{minipage}[t]{0.33\linewidth}
\centering
\includegraphics[width=1.2in]{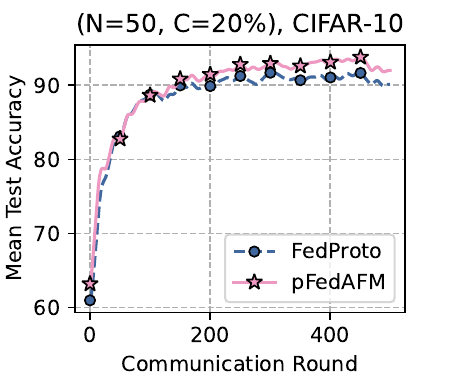}
\end{minipage}%
\begin{minipage}[t]{0.33\linewidth}
\centering
\includegraphics[width=1.2in]{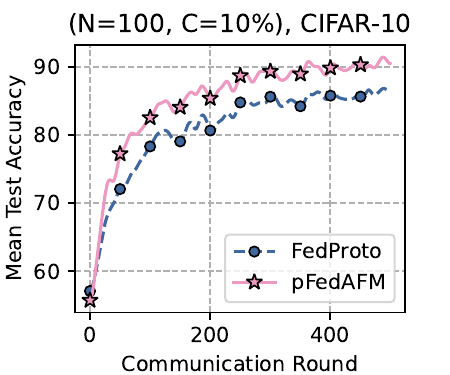}
\end{minipage}%

\begin{minipage}[t]{0.33\linewidth}
\centering
\includegraphics[width=1.2in]{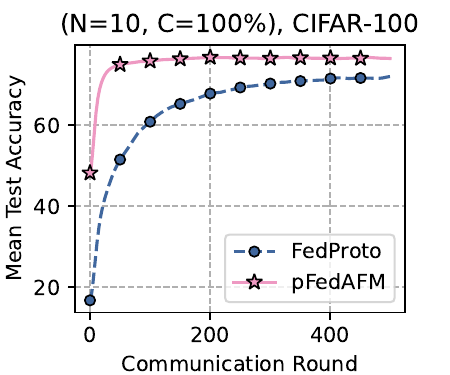}
\end{minipage}%
\begin{minipage}[t]{0.33\linewidth}
\centering
\includegraphics[width=1.2in]{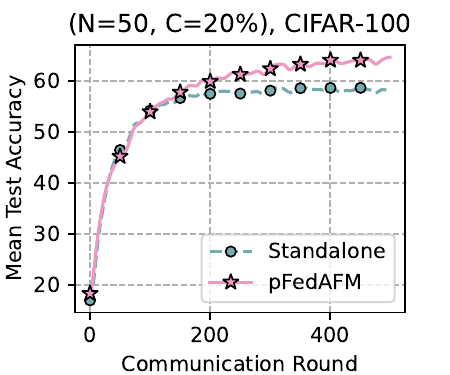}
\end{minipage}%
\begin{minipage}[t]{0.3\linewidth}
\centering
\includegraphics[width=1.2in]{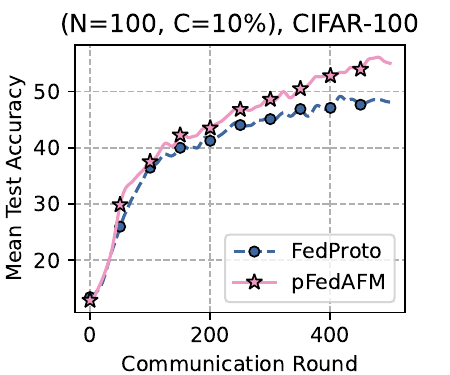}
\end{minipage}%
\caption{Average accuracy varies as rounds. {\tt{Standalone}} or {\tt{FedProto}} is the best baseline in each setting of Table~\ref{tab:compare-hetero}.}
\label{fig:compare-hetero-converge}
\end{figure}

\textbf{Mean Accuracy.}
Table~\ref{tab:compare-hetero} shows that \methodname{} obtains the highest mean accuracy across all FL settings. It achieves up to a $7.93\%$ accuracy improvement over each FL setting's best baseline, and up to $24.39\%$ accuracy increment over each FL setting's best same-category baseline. Significant accuracy improvements demonstrate that \methodname{} with a batch-level \persN based on dynamic dimension-level feature mixture adaptive to local data distribution enhances \genN and \persN of local models. 
Figure~\ref{fig:compare-hetero-converge} shows  
that \methodname{} converges faster to higher accuracy. This reflects that \methodname{} with adaptive batch-level \persN promotes model convergence.

\begin{figure}[t]
\centering
\begin{minipage}[t]{0.5\linewidth}
\centering
\includegraphics[width=1.7in]{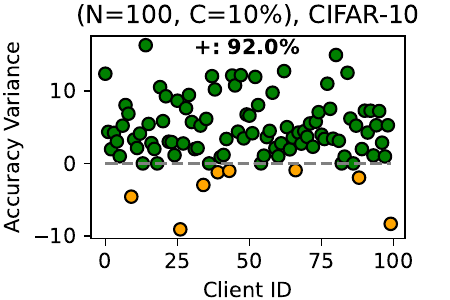}
\end{minipage}%
\begin{minipage}[t]{0.5\linewidth}
\centering
\includegraphics[width=1.7in]{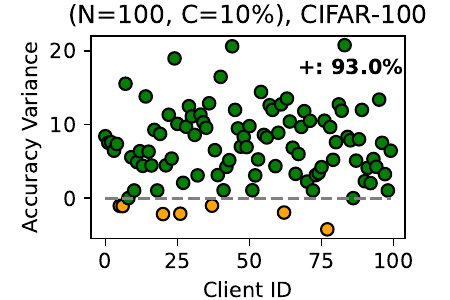}
\end{minipage}%
\caption{Accuracy variance of individual clients.}
\label{fig:compare-individual}
\end{figure}

\textbf{Individual Accuracy.} Figure~\ref{fig:compare-individual} depicts the individual accuracy variance of \methodname{} and the \sota baseline - {\tt{FedProto}} under the most complicated FL setting ($N=100, C=10\%$). 
It can be observed that $92\%$ and $93\%$ of the clients in \methodname{} performs better than those in {\tt{FedProto}} on CIFAR-10 and CIFAR-100, respectively, verifying that \methodname{} with adaptive batch-level \persN enhances the \persN of local models.

\begin{figure}[t]
\centering
\begin{minipage}[t]{0.33\linewidth}
\centering
\includegraphics[width=1.18in]{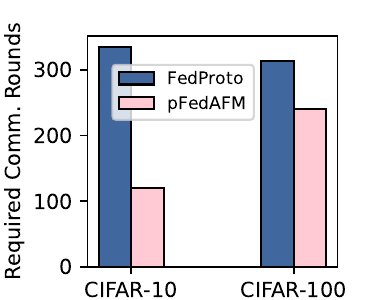}
\end{minipage}%
\begin{minipage}[t]{0.33\linewidth}
\centering
\includegraphics[width=1.18in]{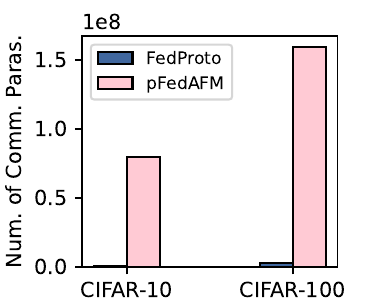}
\end{minipage}%
\begin{minipage}[t]{0.33\linewidth}
\centering
\includegraphics[width=1.18in]{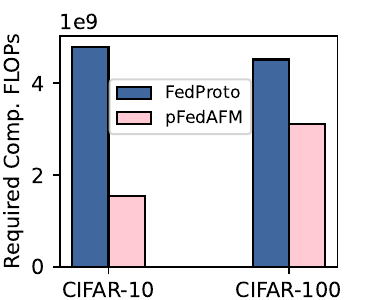}
\end{minipage}%
\caption{Rounds, communication, and computation for target mean accuracy $90\%$ on CIFAR-10 and $50\%$ on CIFAR-100.}
\label{fig:compare-comm-comp}
\end{figure}

\textbf{Communication Cost.} Under ($N=100, C=10\%$), we measure the total communication costs of \methodname{} and {\tt{FedProto}} reaching $90\%$ and $50\%$ target mean accuracy on CIFAR-10 and CIFAR-100, respectively.
Figure~\ref{fig:compare-comm-comp}(middle) shows that \methodname{} incurs a higher total communication cost than {\tt{FedProto}}. Since \methodname{} exchanges \homo feature extractors while {\tt{FedProto}} exchanges local seen-class average representations, \methodname{} incurs a higher one-round communication round, although Figure~\ref{fig:compare-comm-comp}(left) shows that \methodname{} requires fewer rounds to reach the target mean accuracy. Nevertheless, compared with exchanging the complete local models under {\tt{FedAvg}}, \methodname{} still incurs lower communication costs.

\textbf{Computational Overhead.} Figure~\ref{fig:compare-comm-comp}(right) shows that \methodname{} incurs a lower computational cost than {\tt{FedProto}}. {\tt{FedProto}} has to compute the representations of each local sample after training and calculate each seen class's average representation, while \methodname{} only needs to perform model training. Thus, \methodname{} incurs a lower per-round computational cost. Figure~\ref{fig:compare-comm-comp}(left) shows that \methodname{} requires fewer rounds to achieve the target mean accuracy. Thus, it incurs a lower total computational overhead.

\subsection{Case Studies}

\subsubsection{Robustness to Pathological Non-IIDness}
\begin{figure}[t]
\centering
\begin{minipage}[t]{0.5\linewidth}
\centering
\includegraphics[width=1.75in]{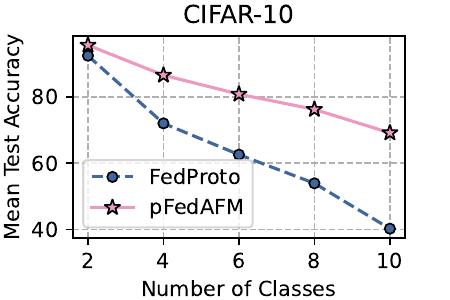}
\end{minipage}%
\begin{minipage}[t]{0.5\linewidth}
\centering
\includegraphics[width=1.75in]{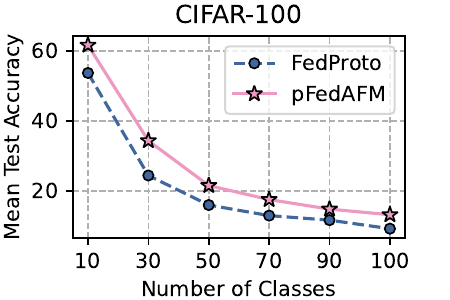}
\end{minipage}%
\caption{Robustness to pathological non-IIDness.}
\label{fig:case-noniid-class}
\end{figure}

We vary the number of seen classes for one client as $\{2, 4, 6, 8, 10\}$ on CIFAR-10 and $\{10, 30, 50, 70, 90, 100\}$ on CIFAR-100 under ($N=100, C=10\%$). Figure~\ref{fig:case-noniid-class} shows that \methodname{} always outperforms {\tt{FedProto}} across all non-IIDness, indicating its robustness to non-IIDness.
The model accuracy of the two algorithms degrades as the non-IIDness drops (the number of seen classes for one client increases). The more seen classes one client holds, the classification ability of the local model to each class drops (\emph{i.e.}, the local model's \genN is enhanced, but \persN is weakened).

\begin{figure}[t]
\centering
\begin{minipage}[t]{0.5\linewidth}
\centering
\includegraphics[width=1.75in]{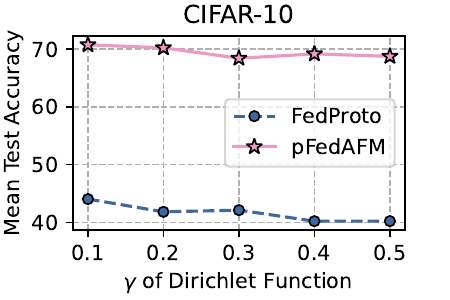}
\end{minipage}%
\begin{minipage}[t]{0.5\linewidth}
\centering
\includegraphics[width=1.75in]{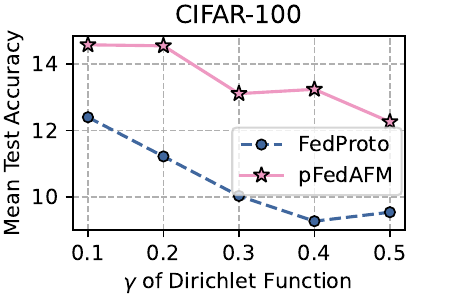}
\end{minipage}%
\vspace{-1em}
\caption{Robustness to practical non-IIDness.}
\label{fig:case-noniid-distribution}
\end{figure}

\subsubsection{Robustness to Practical Non-IIDness}
We vary the hyperparameter $\gamma$ of the Dirichlet function as $\{0.1, 0.2, 0.3, 0.4, 0.5\}$ on two datasets. Figure~\ref{fig:case-noniid-distribution} shows that \methodname{} always performs higher accuracy than {\tt{FedProto}}. Similar to the results under pathological non-IIDness, the model accuracy of the two methods also degrades as the non-IIDness drops ($\gamma$ increases).

\subsubsection{Robustness to Learning Rate of Trainable Weight Vectors}
\methodname{} only introduces one hyperparameter - the learning rate $\eta_{\boldsymbol{\alpha}}$ of trainable weight vectors. We vary $\eta_{\boldsymbol{\alpha}}=\{0.001, 0.01, 0.1, 1\}$ on CIFAR-10 (non-IID:2/10) and CIFAR-100 (non-IID:10/100). For each setting, we use three random seeds for three trials. The results are displayed in Figure~\ref{fig:case-sensitivity-alpha}(left), the dots and shadows denote the average accuracy and its variance.
It can be observed that \methodname{}'s accuracy rises as $\eta_{\boldsymbol{\alpha}}$ increases on CIFAR-10 and no obvious accuracy variations on CIFAR-100. Its accuracy under almost all $\eta_{\boldsymbol{\alpha}}$ settings outperform {\tt{FedProto}} ($92.49\%$ on CIFAR-10, $53.67\%$ on CIFAR-100), indicating its robustness to this hyperparameter.

\subsection{Analysis of Trainable Weight Vectors}\label{sec:alpha-analysis}
We randomly choose $2$ clients under $(N=100, C=10\%)$ on CIFAR-10 and CIFAR-100, respectively. We display the average of each client's trainable weight vector for the local \hetero model varies as rounds.
Figure~\ref{fig:case-sensitivity-alpha}(right) shows that different clients on the same dataset show diverse variations of the trainable weight vector's average values, verifying that clients under \methodname{} train local \pers \hetero models adaptive to local data distributions. Besides, the average weights of the two clients are close to $0$ and $0.5$ on CIFAR-10 and CIFAR-100, respectively, reflecting that the \gen knowledge from the global \homo feature extractor and the \pers knowledge from the local \hetero model contributes differently to model performance in different training tasks. Therefore, it is necessary to utilize a pair of trainable weight vectors to mix the \gen feature and the \pers feature based on local data distributions to balance between model \genN and personalization.

\subsection{Ablation Study}
\subsubsection{Small Model v.s. Large Model v.s. Mixed Model}
\begin{figure}[t]
\centering
\begin{minipage}[t]{0.5\linewidth}
\centering
\includegraphics[width=1.75in]{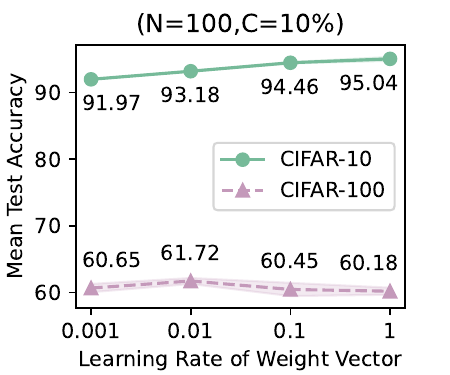}
\end{minipage}%
\begin{minipage}[t]{0.5\linewidth}
\centering
\includegraphics[width=1.75in]{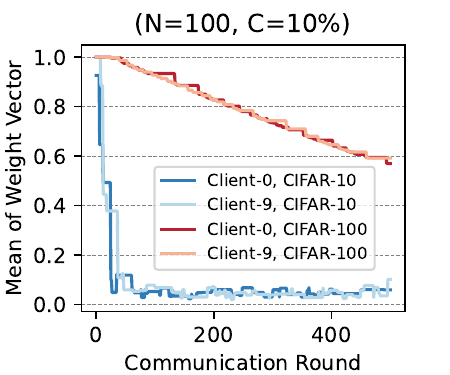}
\end{minipage}%
\caption{Left: robustness to the learning rate of the trainable weight vector $\boldsymbol{\alpha}$. Right: Mean of the trainable weight vector $\boldsymbol{\alpha}$ varies as communication rounds.}
\label{fig:case-sensitivity-alpha}
\vspace{-1em}
\end{figure}

Section~\ref{sec:alpha-analysis} shows that the average of the weight vector for the local \hetero model's representations closes to $0$ on CIFAR-10. We further verify the performance variation among the small model (the global \homo feature extractor and the local header), the large local \hetero model, and the mixed complete model under $(N=100, C=10\%)$. Figure~\ref{fig:ablation}(left) shows that the accuracy values of the three models are from low to high on both datasets, demonstrating the superiority of the mixed complete model. Although the weight vector's average value of the local \hetero model is close to $0$ on CIFAR-10, as shown in Figure~\ref{fig:case-sensitivity-alpha}(right), it still contributes to the performance of the mixed model.

\subsubsection{End-to-End Training v.s. Iterative Training}
\begin{figure}[t]
\centering
\begin{minipage}[t]{0.5\linewidth}
\centering
\includegraphics[width=1.75in]{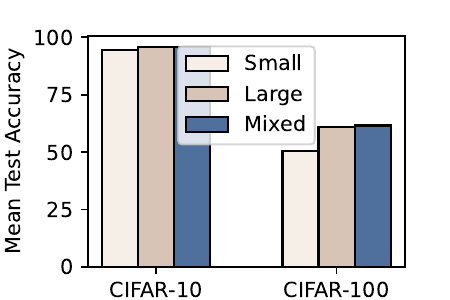}
\end{minipage}%
\begin{minipage}[t]{0.5\linewidth}
\centering
\includegraphics[width=1.75in]{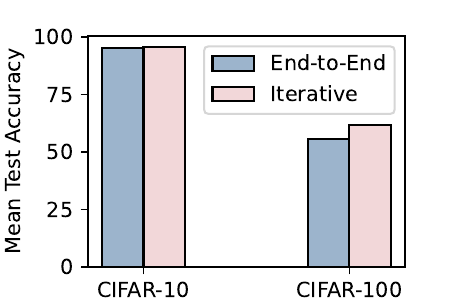}
\end{minipage}%
\caption{Left: mean test accuracy of global homogeneous small model, local heterogeneous large model, and the mixed model. Right: end-to-end training v.s. iterative training.}
\label{fig:ablation}
\end{figure}

An intuitive manner of training the global \homo feature extractor and the local \hetero model is updating them simultaneously in an end-to-end manner. It saves per-round training time compared with the proposed iterative training. Therefore, we study the accuracy achieved by these two approaches under $(N=100, C=10\%)$. Figure~\ref{fig:ablation}(right) shows that iterative training outperforms end-to-end training on both datasets, particularly on the more sophisticated CIFAR-100, verifying that \methodname{} improves model performance via effective bi-directional knowledge transfer.

\section{Discussion}
In this section, we discuss the privacy, communication and computational costs of the proposed \methodname{} as follows.

\textbf{Privacy.} Under \methodname{}, only the small \homo feature extractors are transmitted between clients and the server. The local data and \hetero models are never exposed. Thus, data and model privacy is preserved.

\textbf{Communication Costs.} Clients and the server exchange the small \homo feature extractors which are much smaller than the local \hetero models. The communication costs are lower than exchanging the local \hetero models.

\textbf{Computational Costs.} Besides training the local \hetero models, clients also train the weight vectors and the small \homo feature extractors. Although the computational cost in one round is higher, the total computational costs also depend on the rounds required for convergence. The adaptive batch-level \persN improves convergence rates. Thus, the overall computational costs are acceptable.

\section{Conclusions} 
This paper proposed an effective and efficient MHPFL algorithm, \methodname{}, capable of handling batch-level data heterogeneity. It fuses the knowledge of local \hetero models by sharing an additional \homo small feature extractor across clients. Iterative training effectively facilitates the bidirectional transfer of global \gen knowledge and local \pers knowledge, enhancing the \genN and \persN of local models. The trainable weight vectors dynamically balance the \genN and \persN by mixing \gen features and \pers features adaptive to local data distribution variations across different batches. Theoretical analysis proves it can converge over time with a $\mathcal{O}(1/T)$ convergence rate. Experiments demonstrate that it obtains the \sota model performance with low communication and computational costs.



\bibliographystyle{ACM-Reference-Format}
\bibliography{sample-base}
\appendix
\onecolumn

\section{Theoretical Proofs}\label{app:proof}
\subsection{Proof for Lemma~\ref{lemma:localtraining}}
An arbitrary client $k$'s local mixed complete model $h$ can be updated by $h_{t+1}=h_t-\eta g_{h,t}$ in the (t+1)-th round, and following Assumption~\ref{assump:Lipschitz}, we can obtain
\begin{equation}
\begin{aligned}
 \mathcal{L}_{t E+1} &\leq \mathcal{L}_{t E+0}+\langle\nabla \mathcal{L}_{t E+0},(h_{t E+1}-h_{t E+0})\rangle+\frac{L_1}{2}\|h_{t E+1}-h_{t E+0}\|_2^2 \\
& =\mathcal{L}_{t E+0}-\eta\langle\nabla \mathcal{L}_{t E+0}, g_{h, t E+0}\rangle+\frac{L_1 \eta^2}{2}\|g_{h, t E+0}\|_2^2 .   
\end{aligned}
\end{equation}

Taking the expectation of both sides of the inequality concerning the random variable $\xi_{tE+0}$, we obtain
\begin{equation}
\begin{aligned}
 \mathbb{E}[\mathcal{L}_{t E+1}] &\leq \mathcal{L}_{t E+0}-\eta \mathbb{E}[\langle\nabla \mathcal{L}_{t E+0}, g_{h, t E+0}\rangle]+\frac{L_1 \eta^2}{2} \mathbb{E}[\|g_{h, t E+0}\|_2^2] \\
& \stackrel{(a)}{=} \mathcal{L}_{t E+0}-\eta\|\nabla \mathcal{L}_{t E+0}\|_2^2+\frac{L_1 \eta^2}{2} \mathbb{E}[\|g_{h, t E+0}\|_2^2] \\
& \stackrel{(b)}{\leq} \mathcal{L}_{t E+0}-\eta\|\nabla \mathcal{L}_{t E+0}\|_2^2+\frac{L_1 \eta^2}{2}(\mathbb{E}[\|g_{h, t E+0}\|]_2^2+\operatorname{Var}(g_{h, t E+0})) \\
& \stackrel{(c)}{=} \mathcal{L}_{t E+0}-\eta\|\nabla \mathcal{L}_{t E+0}\|_2^2+\frac{L_1 \eta^2}{2}(\|\nabla \mathcal{L}_{t E+0}\|_2^2+\operatorname{Var}(g_{h, t E+0})) \\
& \stackrel{(d)}{\leq} \mathcal{L}_{t E+0}-\eta\|\nabla \mathcal{L}_{t E+0}\|_2^2+\frac{L_1 \eta^2}{2}(\|\nabla \mathcal{L}_{t E+0}\|_2^2+\sigma^2) \\
& =\mathcal{L}_{t E+0}+(\frac{L_1 \eta^2}{2}-\eta)\|\nabla \mathcal{L}_{t E+0}\|_2^2+\frac{L_1 \eta^2 \sigma^2}{2}.
\end{aligned}
\end{equation}

(a), (c), (d) follow \assum \ref{assump:Unbiased} and (b) follows $Var(x)=\mathbb{E}[x^2]-(\mathbb{E}[x]^2)$.

Taking the expectation of both sides of the inequality for the model $h$ over $E$ iterations, we obtain

\begin{equation}
\mathbb{E}[\mathcal{L}_{t E+1}] \leq \mathcal{L}_{t E+0}+(\frac{L_1 \eta^2}{2}-\eta) \sum_{e=1}^E\|\nabla \mathcal{L}_{t E+e}\|_2^2+\frac{L_1 E \eta^2 \sigma^2}{2} . 
\end{equation}

\subsection{Proof for Lemma~\ref{lemma:aggregation}}
\begin{equation}
\begin{aligned}
\mathcal{L}_{(t+1) E+0}& =\mathcal{L}_{(t+1) E}+\mathcal{L}_{(t+1) E+0}-\mathcal{L}_{(t+1) E} \\
& \stackrel{(a)}{\approx} \mathcal{L}_{(t+1) E}+\eta\|\theta_{(t+1) E+0}-\theta_{(t+1) E}\|_2^2 \\
& \stackrel{(b)}{\leq} \mathcal{L}_{(t+1) E}+\eta \delta^2.
\end{aligned}
\end{equation}

(a): we can use the gradient of parameter variations to approximate the loss variations, \emph{i.e.}, $\Delta\mathcal{L}\approx \eta\cdot \|\Delta \theta\|_2^2$. (b) follows \assum \ref{assump:BoundedVariation}.

Taking the expectation of both sides of the inequality to the random variable $\xi$, we obtain
\begin{equation}
    \mathbb{E}[\mathcal{L}_{(t+1)E+0}]\le\mathbb{E}[\mathcal{L}_{tE+1}]+{\eta\delta}^2.
\end{equation}

\subsection{Proof for Theorem~\ref{theorem:one-round}}
Substituting Lemma~\ref{lemma:localtraining} into the right side of Lemma~\ref{lemma:aggregation}'s inequality, we obtain
\begin{equation}\label{eq:theorem1}
\mathbb{E}[\mathcal{L}_{(t+1) E+0}] \leq \mathcal{L}_{t E+0}+(\frac{L_1 \eta^2}{2}-\eta) \sum_{e=0}^E\|\nabla \mathcal{L}_{t E+e}\|_2^2+\frac{L_1 E \eta^2 \sigma^2}{2}+\eta \delta^2.
\end{equation}

\subsection{Proof for Theorem~\ref{theorem:non-convex}}
Interchanging the left and right sides of Eq.~(\ref{eq:theorem1}), we obtain
\begin{equation}
\sum_{e=0}^E\|\nabla \mathcal{L}_{t E+e}\|_2^2 \leq \frac{\mathcal{L}_{t E+0}-\mathbb{E}[\mathcal{L}_{(t+1) E+0}]+\frac{L_1 E \eta^2 \sigma^2}{2}+\eta \delta^2}{\eta-\frac{L_1 \eta^2}{2}}.
\end{equation}

Taking the expectation of both sides of the inequality over rounds $t= [0, T-1]$ to $W$, we obtain
\begin{equation}
\frac{1}{T} \sum_{t=0}^{T-1} \sum_{e=0}^{E-1}\|\nabla \mathcal{L}_{t E+e}\|_2^2 \leq \frac{\frac{1}{T} \sum_{t=0}^{T-1}[\mathcal{L}_{t E+0}-\mathbb{E}[\mathcal{L}_{(t+1) E+0}]]+\frac{L_1 E \eta^2 \sigma^2}{2}+\eta \delta^2}{\eta-\frac{L_1 \eta^2}{2}}.
\end{equation}

Let $\Delta=\mathcal{L}_{t=0} - \mathcal{L}^* > 0$, then $\sum_{t=0}^{T-1}[\mathcal{L}_{t E+0}-\mathbb{E}[\mathcal{L}_{(t+1) E+0}]] \leq \Delta$, we can get 
\begin{equation}\label{eq:theorem2}
\frac{1}{T} \sum_{t=0}^{T-1} \sum_{e=0}^{E-1}\|\nabla \mathcal{L}_{t E+e}\|_2^2 \leq \frac{\frac{\Delta}{T}+\frac{L_1 E \eta^2 \sigma^2}{2}+\eta \delta^2}{\eta-\frac{L_1 \eta^2}{2}}.
\end{equation}

If the above equation converges to a constant $\epsilon$, \emph{i.e.},

\begin{equation}
\frac{\frac{\Delta}{T}+\frac{L_1 E \eta^2 \sigma^2}{2}+\eta \delta^2}{\eta-\frac{L_1 \eta^2}{2}}<\epsilon,
\end{equation}
then 
\begin{equation}
T>\frac{\Delta}{\epsilon(\eta-\frac{L_1 \eta^2}{2})-\frac{L_1 E \eta^2 \sigma^2}{2}-\eta \delta^2}.
\end{equation}

Since $T>0, \Delta>0$, we can get
\begin{equation}
\epsilon(\eta-\frac{L_1 \eta^2}{2})-\frac{L_1 E \eta^2 \sigma^2}{2}-\eta \delta^2>0.
\end{equation}

Solving the above inequality yields

\begin{equation}
\eta<\frac{2(\epsilon-\delta^2)}{L_1(\epsilon+E \sigma^2)}.
\end{equation}

Since $\epsilon,\ L_1,\ \sigma^2,\ \delta^2$ are all constants greater than 0, $\eta$ has solutions.
Therefore, when the learning rate $\eta$ satisfies the above condition, any client's local mixed complete \hetero model can converge. Notice that the learning rate of the local complete \hetero model involves $\{\eta_\theta,\eta_\omega,\eta_{\boldsymbol{\alpha}}\}$, so it's crucial to set reasonable them to ensure model convergence. Since all terms on the right side of Eq.~(\ref{eq:theorem2}) except for $1/T$ are constants, hence \methodname{}'s non-convex convergence rate is $\epsilon \sim \mathcal{O}(1/T)$.
\end{document}